\documentclass[journal,twocolumn]{IEEEtran}

\usepackage{amsmath}
\usepackage{subfigure}
\usepackage{caption}
\usepackage{algorithm}
\usepackage{algpseudocode}
\usepackage{setspace}
\usepackage{color,hyperref}

\usepackage{graphicx}
\usepackage{amsmath,amssymb} % define this before the line numbering.
\usepackage{epsfig}
\usepackage{amsmath}
\usepackage{amssymb}
\usepackage{colortbl}
\usepackage{soul}
\usepackage{multirow}
\usepackage{pifont}
\newcommand{\xmark}{\ding{55}}

\definecolor{mygray}{gray}{.9}

\newlength\savedwidth
\newcommand\whline{\noalign{\global\savedwidth\arrayrulewidth
		\global\arrayrulewidth 1.25pt}%
	\hline
	\noalign{\global\arrayrulewidth\savedwidth}}

\ifCLASSINFOpdf
\else
\fi

\definecolor{darkblue}{rgb}{0.0,0.0,1.0}
\hypersetup{colorlinks,breaklinks,
	linkcolor=darkblue,urlcolor=darkblue,
	anchorcolor=darkblue,citecolor=darkblue}

\begin{document}
\setul{}{1.5pt}

\title{Weakly Supervised Adversarial Domain Adaptation \\ for Semantic Segmentation in Urban Scenes}

\author{Qi~Wang,\IEEEmembership{~Senior Member,~IEEE}, Junyu~Gao, and ~Xuelong~Li,\IEEEmembership{~Fellow,~IEEE}
	% <-this % stops a space
	\thanks{		
		This work was supported by the National Natural Science Foundation of China under Grant U1864204 and 61773316, Natural Science Foundation of Shaanxi Province under Grant 2018KJXX-024, and Project of Special Zone for National Defense Science and Technology Innovation.
		
		Qi Wang, Junyu Gao and Xuelong Li are with the School of Computer Science and the Center for
		OPTical IMagery Analysis and Learning (OPTIMAL), Northwestern Polytechnical
		University, Xi'an 710072, China (e-mail: crabwq@gmail.com; gjy3035@gmail.com; xuelong\_li@nwpu.edu.cn).
		
		\copyright 20XX IEEE. Personal use of this material is permitted. Permission from IEEE must be obtained for all other uses, in any current or future media, including reprinting/republishing this material for advertising or promotional	purposes, creating new collective works, for resale or redistribution to servers or lists, or reuse of any copyrighted component of this work in other works.}% <-this % stops a space
	%\thanks{Manuscript received April 19, 2005; revised January 11, 2007.}}
}
\markboth{{IEEE} Transactions on Image Processing}%
{Shell \MakeLowercase{\textit{et al.}}: Bare Demo of IEEEtran.cls for Journals}
%make the title area
\maketitle

\begin{abstract}
Semantic segmentation, a pixel-level vision task, is developed rapidly by using convolutional neural networks (CNNs). Training CNNs requires a large amount of labeled data, but manually annotating data is difficult. For emancipating manpower, in recent years, some synthetic datasets are released. However, they are still different from real scenes, which causes that training a model on the synthetic data (source domain) cannot achieve a good performance on real urban scenes (target domain). In this paper, we propose a weakly supervised adversarial domain adaptation to improve the segmentation performance from synthetic data to real scenes, which consists of three deep neural networks. To be specific, a detection and segmentation (``DS'' for short) model focuses on detecting objects and predicting segmentation map; a pixel-level domain classifier (``PDC'' for short) tries to distinguish the image features from which domains; an object-level domain classifier (``ODC'' for short) discriminates the objects from which domains and predicts the objects classes. PDC and ODC are treated as the discriminators, and DS is considered as the generator. By the adversarial learning, DS is supposed to learn domain-invariant features. In experiments, our proposed method yields the new record of mIoU metric in the same problem.
	
\end{abstract}

\section{Introduction}
Semantic segmentation is a fundamental task in computer vision, which is viewed as a union of image segmentation, object localization and multi-object recognition. For the specific scenes (such as urban and indoor scenes), the task can be named as fully scene labeling/parsing, which requires to predict the label for each pixel. This paper will focus on the fully urban scenes labeling.

%Scene labeling is a fundamental task in computer vision, which is viewed as a union of image segmentation, object localization and multi-object recognition. Some researchers usually treat it as semantic segmentation, but it is slightly different from semantic segmentation. Scene labeling requires to predict the label for each pixel while semantic segmentation only aims to segment the foreground objects. This paper will focus on the urban scenes labeling.

Recently, convolutional neural networks (CNNs) have obtained the amazing performances in the three fundamental vision tasks: image classification \cite{krizhevsky2012imagenet,simonyan2014very,he2016deep}, object detection \cite{ren2015faster,liu2016ssd}, and semantic segmentation \cite{DBLP:journals/corr/LongSD14}. However, training CNNs requires a large amount of labeled data. Especially, for the scene labeling, annotating images for each pixel is more difficult and expensive than the other two tasks. Thus, the current pixel-wise urban datasets (such as CamVid \cite{DBLP:journals/prl/BrostowFC09} and Cityscapes \cite{Cordts2016Cityscapes}) contain no more than 10,000 images, which is insufficient for some practical applications (e.g. self-driving cars). 

In order to address the data shortage problem, some weakly supervised methods \cite{papandreou2015weakly,oh2017exploiting} try to segment the image by exploiting some weak labels (image-level or object-level labels). However, they only focus on the salient foreground objects segmentation in simple scenes. In the urban scenes, the above methods cannot effectively learn discriminative features from the weakly labels because of many objects with different scales and occlusion, especially background objects (such as road, sky, building and so on). To the best of our knowledge, no algorithm tackles the labeling of full scenes via the weakly supervised learning.

In addition to the strategy at the methodology level, a potential idea is to exploit the synthetic data to prompt the performance in the real world. In recent years, some large-scale synthetic datasets \cite{Richter_2016_ECCV,RosCVPR16} are released, which are generated by computer graphics or crawled from some computer games. The emergence of synthetic datasets greatly emancipates manpower. Unfortunately, there exist significant domain gaps between the synthetic images and real images, including image textures, architectural styles, road materials and so on. As a result, it leads to poor performances when applying the model trained on synthetic images to real scenes. This phenomenon shows that existing supervised strategies may over learn the local discriminative features in the given training data space. 

\begin{figure}[t]
	\centering
	\includegraphics[width=.5\textwidth]{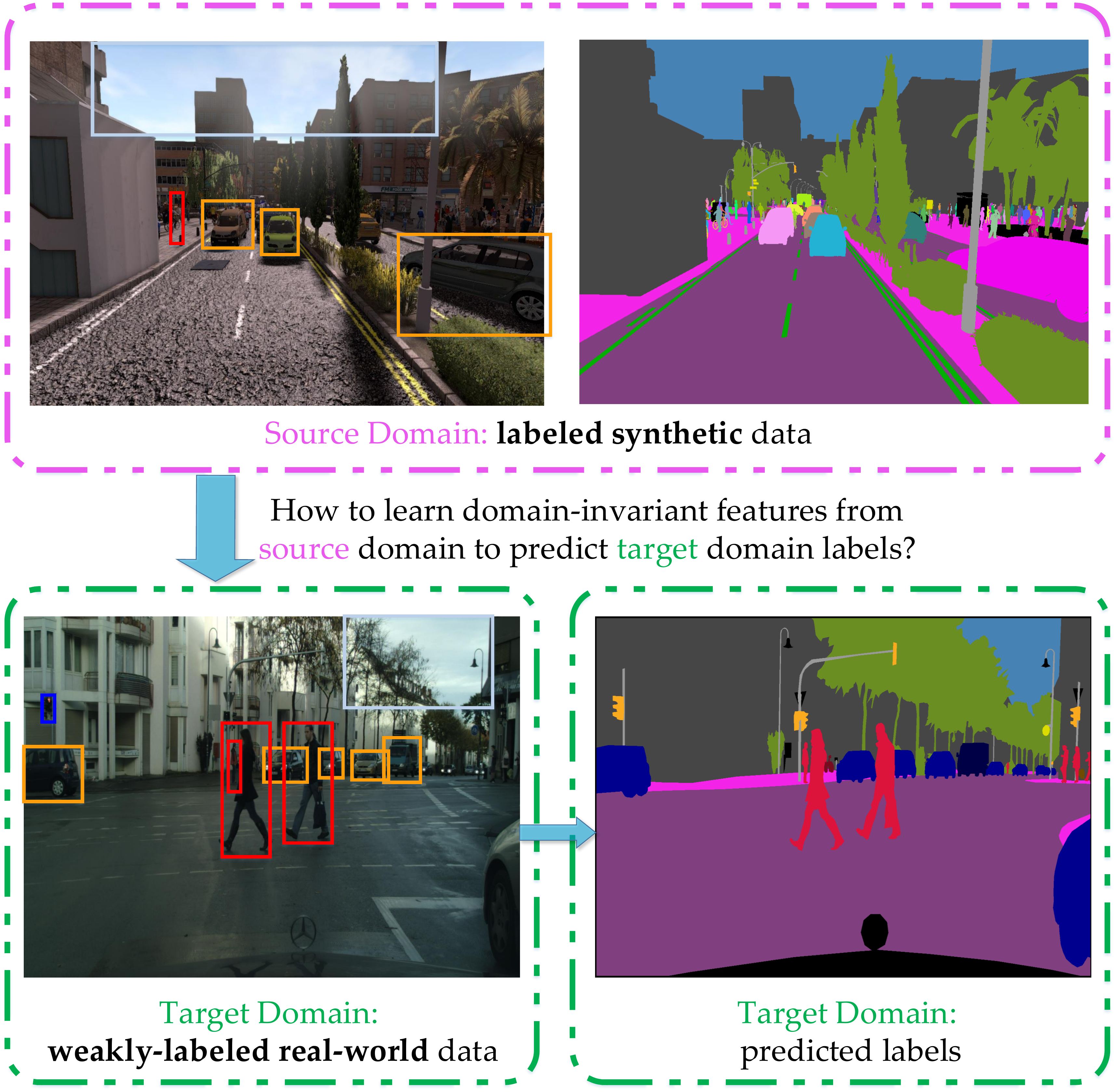}
	\caption{Weakly supervised domain adaptation approach for semantic segmentation in real urban scenes. Given a source domain (synthetic data) with pixel/object- level labels, and a target domain (real-world scenes) with only object-level labels, our goal is train a segmentation model to predict the per-pixel labels of the target domain. }\label{Fig-show}
\end{figure}

The above cross-domain (from the synthetic data to the real-world scenes) semantic segmentation attracts many researchers' attentions. There are two unsupervised FCN-based domain adaptation methods \cite{hoffman2016fcns,zhang2017curriculum} to address the cross-domain problem. However, they only focus on the local pixel-level features while ignore structured object-level features in the scenes. As a matter of fact, some object-level features in the synthetic scenes are similar to that in real urban scenes, which are more robust than the pixel-level features for the cross-domain task. In general, the cross-domain generalization ability of object detection model are stronger than that of segmentation models.

Motivated by the above observation and some recent adversarial learning works and unsupervised methods \mbox{\cite{tzeng2017adversarial,DBLP:conf/icml/LongZ0J17,wang2018detecting,wang2018spectral}}, in this paper, a weakly supervised adversarial domain adaptation approach is proposed to improve the segmentation performance from synthetic data (source domain) to real scenes (target domain). Figure \ref{Fig-show} briefly shows the problem setting: the source domain needs to provide the pixel-level and object-level labels, and the target domain only provides the object-level labels.

\begin{figure*}[t]
	\centering
	\includegraphics[width=0.98\textwidth]{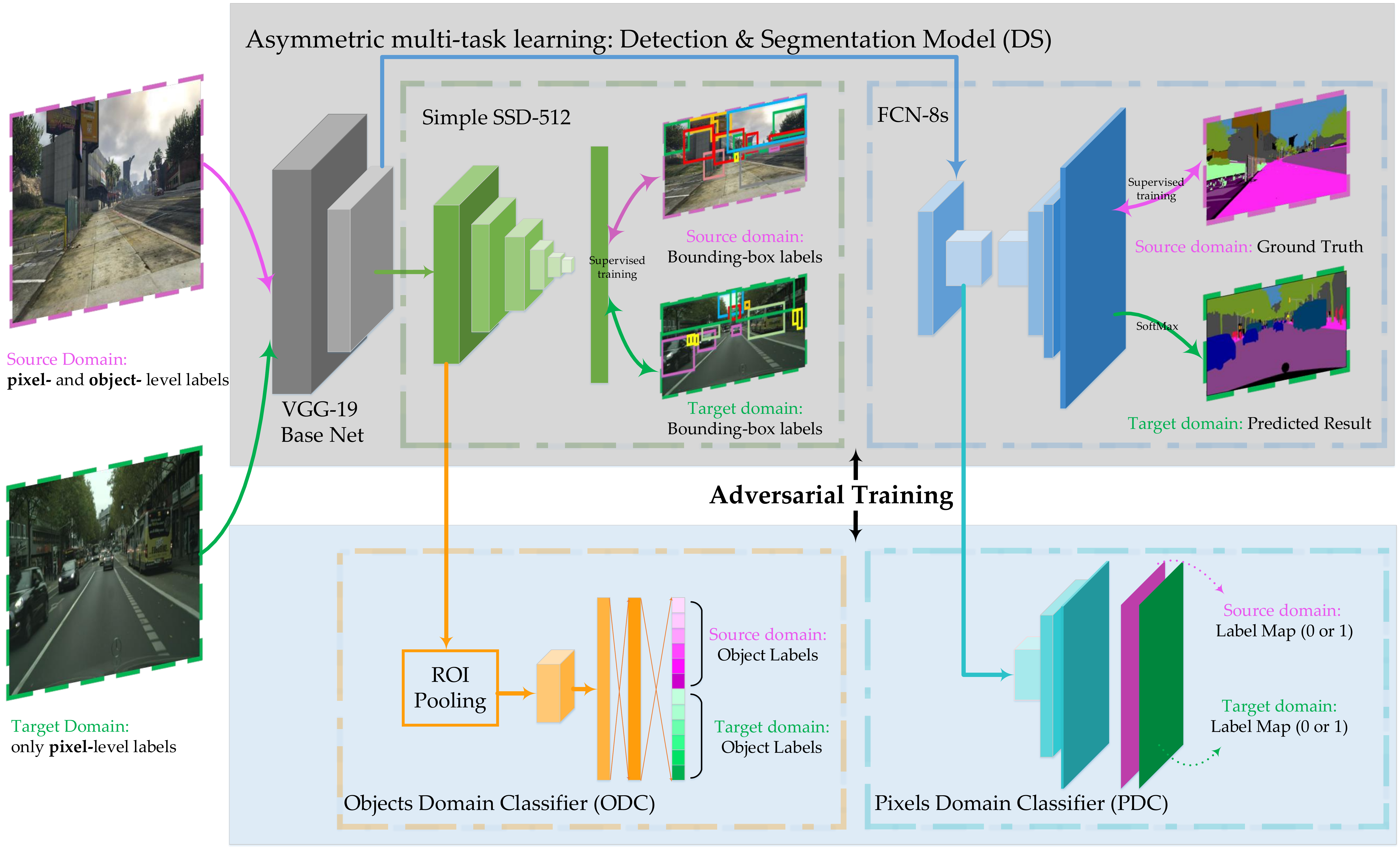}
	\caption{The flowchart of the proposed weakly supervised adversarial domain adaptation. On the top, the asymmetric multi-task model is depicted, which consists of a detection model and a segmentation model (DS). During the training stage, a pair of images from two domains are fed to the DS model. The {\color{magenta}{magenta}} and {\color{green}{green}} curve arrows represent the input/output of source and target domain, respectively. Further, the two-way arrow shows that the data flow is involved in the training process. From this figure, source images take part in the object- and pixel- level training, while target images only participate in the object-level training. On the bottom, the two domain classifiers (PDC and ODC) at the object- and pixel- levels are demonstrated. The feature maps of two streams in DS are respectively fed to PDC and ODC, respectively. By alternately adversarial optimizing DS and two domain classifiers, the final DS will be obtained. During the testing phase, the test images are only fed to the segmentation stream in DS to predict the pixel-level score map.}\label{Fig-overview}
\end{figure*}

Figure \ref{Fig-overview} illustrates the entire framework. To be specific, the proposed method consists of three deep neural networks, a multi-task model for object Detection and semantic Segmentation (DS), a Pixel-level Domain Classifier (PDC) and an Object-level Domain Classifier (ODC). DS integrates a detection network and a segmentation network into one architecture. The former focuses on learning object-level features to localize the objects' bounding boxes, and the latter aims to learn local features to classify each pixel. PDC is fed with the feature maps of the segmentation network, and outputs their domain (source or target domain) for each pixel. ODC is fed with the objects features of detection network, then outputs objects category and domain class. Similar to the generative adversarial learning \cite{goodfellow2014generative}, DS model can be treated as a generator, and PDC/ODC models are regarded as two discriminators. After the adversarial training, DS model can learn domain-invariant features at the pixel and object levels to confuse PDC and ODC.

In summary, the main contributions of this paper are:
\begin{enumerate}
	\item[1)] To our best knowledge, this paper is one of the first attempts to propose a weakly supervised method for fully urban scenes labeling, which employs the cross-domain problem. It can extract more robust domain-invariant features than the traditional FCN-based methods.
	
	\item[2)] This paper designs two domain classifiers at the pixel/object levels to distinguish which domain the image features come from. By adversarial training, the domain gap can be effectively reduced.
	
	\item[3)] The proposed method yields a new record of mIoU accuracy on the cross-domain fully urban scenes labeling.  
\end{enumerate}

\section{Related Work}
In this section, we briefly review the important works about the two most related tasks: fully/weakly supervised semantic segmentation, domain adaptation with deep leaning.   

\textbf{Semantic segmentation.} In 2014, fully convolutional network (FCN) proposed by Long \emph{et al.} \mbox{\cite{DBLP:journals/corr/LongSD14}} achieves a significant improvement in the field of some pixel-wise tasks (such as semantic segmentation, saliency detection, crowd density estimation and so on), which is a fully supervised method. After that, more and more methods \cite{zheng2015conditional,7803544,ronneberger2015u,yu2015multi,zhao2016pyramid,gao2018journal_embedding} based on FCNs are presented. Zheng \emph{et al.} \cite{zheng2015conditional} propose an interpretation of dense conditional random fields as recurrent neural networks, which is appended to the top of FCN. Seg-net \cite{7803544} and U-net \cite{ronneberger2015u} develop a symmetrical encoder-decoder architecture to prompt the performance output maps. Yu and Loltum \cite{yu2015multi} propose a dilated convolution operation to aggregate multi-scale contextual information. Zhao \emph{et al.} \cite{zhao2016pyramid} design a pyramid pooling module in FCN to exploit the capability of global context information. He \emph{et al.} \cite{he2017mask} propose a supervised multi-task learning for instance segmentation, which does not segment the background objects. Wang \emph{et al.} \cite{gao2018journal_embedding} present a FCN to combine RGB images and contour information for road region segmentation.

Recently, some weakly-supervised methods \cite{papandreou2015weakly,7775087,jin2017webly,souly2017semi,oh2017exploiting} are presented to save the costs of annotating ground truth. Papandreou \emph{et al.} \cite{papandreou2015weakly} adopt on-line EM (Expectation-Maximization) methods training segmentation model from image-level and bounding-box labels. \cite{7775087,jin2017webly} apply a progressively learning strategy to train DCNN from the image-level images. Souly \emph{et al.} \cite{souly2017semi} apply a Generative Adversarial Networks (GANs) in which a generator network provides extra training data to a classifier. Oh \emph{et al.} \cite{oh2017exploiting} exploit the saliency features as additional knowledge and mine prior information on the object extent and image statistics to segment the object regions.  It is noted that the above mentioned weakly-supervised methods do not focus on labeling of full scenes. They aim to segment the salient foreground objects in the simple scenes.

\textbf{Domain adaptation.} There are two main streams to study domain adaptation. Some methods \cite{tzeng2015simultaneous,ganin2015unsupervised,ganin2016domain,ghifary2016deep,tzeng2017adversarial} attempt to minimize the domain gap via adversarial training. \cite{tzeng2015simultaneous,ganin2015unsupervised,ganin2016domain} propose a Domain-Adversarial Neural Network, which minimizes the domain classification loss. Muhammad \emph{et al.} \cite{ghifary2016deep} propose an DRCN to reconstruct target domain images by optimizing a domain classifier. Tzeng \emph{et al.} \cite{tzeng2017adversarial} present a generalized framework for adversarial adaptation, which help us understand the benefits and key ideas from GANs-based methods. 

Other methods \cite{tzeng2014deep,long2015learning,bousmalis2016domain,DBLP:conf/icml/LongZ0J17} adopt the Maximum Mean Discrepancy (MMD) \cite{gretton2012kernel} to alleviate domain shift. MMD measures the difference between features extracted from each domain. Tzeng \emph{et al.} \cite{tzeng2014deep} computes the MMD loss at one layer and Long \emph{et al.} \cite{long2015learning} minimizing MMD losses at multi-layer Deep Adaptation Network. Bousmalis \emph{et al.} \cite{bousmalis2016domain} propose a Domain Separation Networks (DSN) to learn domain-invariant features by explicitly separating representations private to each domain. Further, Long \emph{et al.} \cite{DBLP:conf/icml/LongZ0J17} combines Joint Adaptation Networks (JAN) with adversarial training strategy.

\textbf{Domain adaptation for semantic segmentation.} Hoffman \emph{et al.} \cite{hoffman2016fcns} firstly propose an unsupervised domain adaptation for segmentation, which combines global and category adaptation in the adversarial learning. It effectively reduces the domain gap at the pixel level. Zhang \emph{et al.} \cite{zhang2017curriculum} adopt a curriculum-style domain adaption and predict global and local label distributions at image and superpixel levels, respectively.

\section{Approach}

This section describes the detailed methodology of the proposed weakly supervised adversarial domain adaptation for semantic segmentation. In order to reduce the domain gap, the inter- and intra- object features are considered in the neural network. In addition, by alternately adversarial optimizing DS and two domain classifiers (PDC and ODC), the domain gap of learned features by DS can be alleviated effectively. Figure \ref{Fig-overview} illustrates the entire framework.

Before the detailed description, it is necessary to recall our faced cross-domain semantic segmentation by mathematical notations. A source domain ${\mathcal{S}}$ from a synthetic urban dataset provides images ${I_\mathcal{S}}$, pixel-level annotations $A_\mathcal{S}^{pix}$, and object-level annotations $A_\mathcal{S}^{obj}$; and a target domain ${\mathcal{T}}$ from real world provides images ${I_\mathcal{T}}$, only object-level annotations $A_\mathcal{T}^{obj}$. Note that ${\mathcal{S}}$ and ${\mathcal{T}}$ share the same label space ${\mathbb{R}^C}$, where ${C}$ is the number of categories. In a word, given ${I_\mathcal{S}}$,  $A_\mathcal{S}^{pix}$, $A_\mathcal{S}^{obj}$, ${I_\mathcal{T}}$ and $A_\mathcal{T}^{obj}$, the goal is to train a segmentation model to predict pixel-wise score map of ${\mathcal{T}}$. 

Under the above definitions, the purpose of this paper is that how to reduce the domain gap between ${\mathcal{S}}$ and ${\mathcal{T}}$.

\subsection{Weak supervision for segmentation}

\label{DS}
Almost all of deep methods for semantic segmentation are based on FCN owing to its powerful learning ability. However, FCN-based methods perform not well for our faced cross-domain problems. The main reasons are that semantic segmentation is considered as a pixel-wise classification problem, and many FCN-based methods focus on the local features (texture, color and so on) and ignore large-scale structured features. Unfortunately, the differences of texture, color or other local features are obvious in the different domains. On the contrary, the structured features and the contextual information are consistent with different domains, for instance, pedestrian posture, vehicle appearance and the position relation of objects. Thus, it is important to extract object-level features for cross-domain semantic segmentation. 

Previous works \cite{hariharan2014simultaneous,he2017mask} tackle object detection and segmentation simultaneously in a single framework. However, as for the target domain with only bounding-box labels, the above supervised methods are impracticable. In this work, we propose an asymmetric multi-task learning to handle it, which consists of Detection and Segmentation streams (DS). During the training stage, a pair of images from two domains are fed to the neural network: the source images are involved in the entire model's training; the target images only participate in the detection stream's training. At the testing phase, the test images are only fed to the segmentation stream in DS to predict the pixel-wise score map. Compared with Mask RCNN \cite{he2017mask}, our model consists of two streams (shown in Fig. \ref{Fig-overview}), which is an asymmetric multi-task learning on the two domains. But Mask RCNN \cite{he2017mask} must detect the objects first and then segment them. In other word, the detection result of Mask RCNN is essential while ours is auxiliary in the test stage. 

To be specific, an FCN-8s \mbox{\cite{DBLP:journals/corr/LongSD14}} is combined with a simple SSD-512 \mbox{\cite{liu2016ssd}} into one architecture, in which the first four groups of convolutional layers are shared (named as \emph{Base Net}). The FCN-8s aims to localize the objects' boundaries and per-pixel segmentation, and the SSD-512 focuses on learning object-level features to localize the objects' bounding boxes. Unlike the traditional detection methods, our SSD-512 can learn not only the structured objects (such as pedestrian, car, bicycle, and so on) but also some unstructured objects (e.g., road, sky, building, etc.). For the structured feature, it is an internal feature of a single object. For example, usually, the pedestrian has one head, two arms, two legs and so on, and these parts present a certain position distribution. Similarly, other objects (cars, truck, traffic sign/light) have specific structured features.About the these large unstructured objects, they contain more contextual information, which is a type of intra-object features. For example, the building is usually located under the sky in urban images, and the rectangular road region may cover the part of vehicles, pedestrians, sidewalks. Similar object relations can be regarded as a type of inter-object feature. 

% todo：Fig. xxx show the exemplars of inter-object relations.

The proposed DS model is trained through following loss:

\begin{equation}
\begin{array}{l}
\begin{aligned}
{\mathcal{L}_{DS}} =& {\mathcal{L}_{seg}}({I_\mathcal{S}},A_\mathcal{S}^{pix}) \\
&+ {\mathcal{L}_{det }}({I_\mathcal{S}},A_\mathcal{S}^{obj}) +{L_{{det} }}({I_\mathcal{T}},A_\mathcal{T}^{obj}),
\end{aligned}
\end{array}
\end{equation}
where ${\mathcal{L}_{seg}}({I_\mathcal{S}},A_\mathcal{S}^{pix})$ is 2D Cross Entropy Loss, the standard supervised pixel-wise classification objective. ${\mathcal{L}_{det }}({I_\mathcal{S}},A_\mathcal{S}^{obj})$ and ${L_{{det} }}({I_\mathcal{T}},A_\mathcal{T}^{obj})$ are MultiBox objective loss functions \cite{liu2016ssd} for the detection task, which is a weighted sum of the localization loss and the confidence loss. 

%To be specific, ${\mathcal{L}_{seg}}({I_\mathcal{S}},A_\mathcal{S}^{pix})$ is formulated as below:
%And ${\mathcal{L}_{det }}({I_\mathcal{S}},A_\mathcal{S}^{obj})$ and ${L_{{det} }}({I_\mathcal{T}},A_\mathcal{T}^{obj})$ is defines as:

\subsection{Adversarial domain adaptation}

Although the proposed weak supervision learns some domain-invariant features (including the structured intra-object feature and the contextual inter-object feature), other domain gap (such as texture, color and so on) is still not alleviated. These differences between synthetic and real-world domains are inherent. For the traditional supervised deep learning, the trained model only learns the discriminative features according to given labeled synthetic data. However, there is a problem that the learned discriminative features are not universal for real-world data. 

Adversarial learning \cite{tzeng2017adversarial} provides a good framework to tackle the above problem, which pits two networks against each other. On the one hand, a domain classifier is trained to distinguish which domain the learned features are from. On the other hand, the original main model is supposed to learn not only the discriminative features to label scenes but also the domain-invariant features to confuse the domain classifier. By alternately training the two models, the extracted features from main model are invariant with respect to the domain gap.

In this paper, the Pixel-level and Object-level Domain Classifiers (PDC and ODC) are designed as the discriminators, and DS is treated as the generator in the GAN theory. Through the adversarial training, DS is supposed to learn domain-invariant features to confuse PDC and ODC. 

\subsection{Pixel-level adaptation}
\label{PDC_al}
Since basic labeling unit of semantic segmentation is the pixel, correspondingly, a pixel-level domain classifier (PDC) is built to distinguish domain source (source domain or target domain) for each pixel. It receives the feature inputs from the segmentation stream in DS and outputs 2-channel score map with the original image's size to represent the confidence scores of per-pixel domain classes. To be specific, it consists of a convolutional layer and two de-convolutional layers. The bottom-right sub figure in Fig. \ref{Fig-overview} shows the network architecture of PDC.

Given the feature input, the PDC loss is computed as follows: 

\begin{equation}
\begin{array}{l}
\begin{aligned}
{\mathcal{L}_{PDC}} = & - \sum\limits_{O_\mathcal{S}^{seg} \in \mathcal{S}} {\sum\limits_{h \in H} {\sum\limits_{w \in W} {\log (p(O_\mathcal{S}^{PDC})) } } } \\
&- \sum\limits_{O_\mathcal{T}^{seg} \in \mathcal{T}} {\sum\limits_{h \in H} {\sum\limits_{w \in W} {\log (1 - p(O_\mathcal{T}^{PDC}))} } } ,
\end{aligned}\label{PDC}
\end{array}
\end{equation}
where $O_\mathcal{S}^{PDC}$ and $O_\mathcal{T}^{PDC}$ are pixel-wise 2D-channel score map with size of $H \times W$ for source and target feature inputs, $H$ and $W$ denote the height and width of the original image, and $p( \cdot )$ is the soft-max operation for each pixel.

At the same time, here, the inverse of PDC loss, ${\mathcal{L}_{PDC_{inv}}}$ is defined as:
\begin{equation}
\begin{array}{l}
\begin{aligned}
{\mathcal{L}_{PDC_{inv}}} = & - \sum\limits_{O_\mathcal{S}^{seg} \in \mathcal{S}} {\sum\limits_{h \in H} {\sum\limits_{w \in W} {\log (1 - p(O_\mathcal{S}^{PDC})) } } } \\
&- \sum\limits_{O_\mathcal{T}^{seg} \in \mathcal{T}} {\sum\limits_{h \in H} {\sum\limits_{w \in W} {\log (p(O_\mathcal{T}^{PDC}))} } }.
\end{aligned}\label{ivsPDC}
\end{array}
\end{equation}

However, optimizing Eq. (\ref{PDC}) and Eq. (\ref{ivsPDC}) are prone to oscillation. In fact, during the practical training phase, a domain confusion objective \cite{tzeng2015simultaneous} is adopted to replace Eq. (\ref{ivsPDC}), which is defined as below:

\begin{equation}
\begin{array}{l}
\begin{aligned}
{\mathcal{\hat{L}}_{PDC_{inv}}} = \dfrac{1}{2}({\mathcal{L}_{PDC}} + {\mathcal{L}_{PDC_{inv}}})
\end{aligned}\label{tr_PDC}.
\end{array}
\end{equation}
Finally, the objectives are written as follows:
\begin{equation}
\begin{array}{l}
\begin{aligned}
\mathop {min}\limits_{{\theta _{PDC}}} \quad {\mathcal{L}_{PDC}},
\end{aligned}\label{loss1-1}
\end{array}
\end{equation}\begin{equation}
\begin{array}{l}
\begin{aligned}
\mathop {min}\limits_{{\theta _{DS}}} \quad {\mathcal{L}_{DS}} + {\hat {\mathcal{L}}}_{PD{C_{inv}}},
\end{aligned}\label{loss1-2}
\end{array}
\end{equation}
where $\theta _{PDC}$ and $\theta _{DS}$ denote the network parameters of PDC and DS, respectively. During the training stage, the parameters of the two models are updated in turns by minimizing Eq. (\ref{loss1-1}) and Eq. (\ref{loss1-2}). To be specific, a) fix $\theta _{DS}$, and update $\theta _{PDC}$ by optimizing Eq. (\mbox{\ref{loss1-1}}); b) fix $\theta _{PDC}$, and update $\theta _{DS}$ by optimizing Eq. (\mbox{\ref{loss1-2}}).

\subsection{Object-level adaptation}

In Section \ref{DS}, the object detection task is introduced in the segmentation network. Naturally, we also think modeling an object-level domain classifier (ODC) is important to extract domain-invariant features. The goal of ODC is distinguishing the object features belong to which category and come from which domain. As for some traditional domain classifiers, they only need to distinguish the data source. Here, the proposed ODC can classify the objects class, which also guides the SSD-512 can more easily learn discriminative object features.

For getting the accurately object features from the feature maps of input images, the ROI (region of interest) pooling operation \cite{girshick2015fast} is a good choice. Note that the location information in ROI pooling is provided by the ground truth. In SSD-512, the filters of different layers are sensitive to the objects with different scales. Especially, the several top layers' spatial outputs are very small ($16 \times 16$, $8 \times 8$, $4 \times 4$ and $2 \times 2$) so that ROI pooling cannot accurately extract the object features. Thus, we select the feature map with $H \times W$ of $32 \times 32$ to extract the object features. 

After the ROI pooling, object features with the same size are fed to ODC, which is a simple classification network. In order to classify the category and domain simultaneously, the last feature vector is mapped into a $2 \times N$-D confidence vector by the linear operation. $N$ is the number of object classes. The items of ${\rm{1}} \sim {\rm{N}}$ and ${\rm{(N + 1)}} \sim {\rm{N*2}}$ in the confidence vector represent the scores of N classes in source domain and target domain, respectively. The bottom-left sub figure in Figure \ref{Fig-overview} describes the network design of ODC.

In ODC, each label is a one-hot vector. For the clearer expression of each label, it is necessary to formulate the one-hot vector. As for the $N$-D one-hot vector $Y_N^{}(c) = [{y_1},{y_2},...,{y_N}]$, each component is defined as follows:

\begin{equation}
\begin{array}{l}
\begin{aligned}
{y_i} = \left\{ \begin{array}{l}
1,\quad if\,i = c\\
0,\quad otherwise
\end{array} \right..
\end{aligned}\label{onehot}
\end{array}
\end{equation}

Then, the labels definitions are reported in ODC as below. To be specific, for an object with class $c$ from the source domain, a one-hot vector $A_\mathcal{S}^c = {Y_{2N}}(c)$ is generated as the label. Similarly, the label of target domain is $A_\mathcal{T}^c = {Y_{2N}}(N+c)$. Finally, our goal is optimizing the ODC loss as below:

\begin{equation}
\begin{array}{l}
\begin{aligned}
{\mathcal{L}_{ODC}} =& CEL(p(O_\mathcal{S}^{ODC}),A_\mathcal{S}^c)\\& + CEL(p(O_\mathcal{T}^{ODC}),A_\mathcal{T}^c),
\end{aligned}\label{loss_ODC}
\end{array}
\end{equation}
where $O_\mathcal{S}^{ODC}$ and $O_\mathcal{T}^{ODC}$ denote the score vector for each object feature, $p( \cdot )$ is the soft-max operation for each pixel, $CEL$ function is the standard Cross Entropy Loss.

At the same time, the inverse of ODC loss should be computed to guide SSD-512 to learn domain-invariant features. To be specific, the inverse labels of the both are defined as follows: $A_{\mathcal{S}_{inv}}^c = {Y_{2N}}(N+c)$ and $A_{\mathcal{T}_{inv}}^c = {Y_{2N}}(c)$, and the inverse of ODC loss, ${\mathcal{L}_{ODC_{inv}}}$ is defined as:
\begin{equation}
\begin{array}{l}
\begin{aligned}
{\mathcal{L}_{ODC_{inv}}} =& CEL(p(O_\mathcal{S}^{ODC}),A_{\mathcal{S}_{inv}}^c)\\& + CEL(p(O_\mathcal{T}^{ODC}),A_{\mathcal{T}_{inv}}^c).
\end{aligned}\label{loss_inv_ODC}
\end{array}
\end{equation}
In order to avoid the oscillation, the domain confusion objective similar to Eq. (\ref{tr_PDC}) are used:

\begin{equation}
\begin{array}{l}
\begin{aligned}
{\mathcal{\hat{L}}_{ODC_{inv}}} = \dfrac{1}{2}({\mathcal{L}_{ODC}} + {\mathcal{L}_{ODC_{inv}}})
\end{aligned}\label{tr_ODC}.
\end{array}
\end{equation}
Given Eq. (\ref{loss_ODC}) and Eq. (\ref{tr_ODC}), similar to Section \ref{PDC_al}, by iteratively optimizing ODC and DS, the final DS is obtained.

Overall, for the full models (including DS, PDC and ODC) training, the objectives are written as follows:
\begin{equation}
\begin{array}{l}
\begin{aligned}
\mathop {min}\limits_{{\theta _{PDC}}} \quad {\mathcal{L}_{PDC}},
\end{aligned}\label{loss1}
\end{array}
\end{equation}\begin{equation}
\begin{array}{l}
\begin{aligned}
\mathop {min}\limits_{{\theta _{ODC}}} \quad {\mathcal{L}_{ODC}},
\end{aligned}\label{loss2}
\end{array}
\end{equation}\begin{equation}
\begin{array}{l}
\begin{aligned}
\mathop {min}\limits_{{\theta _{DS}}} \quad {\mathcal{L}_{DS}} + {\hat {\mathcal{L}}}_{PD{C_{inv}}} + {\hat {\mathcal{L}}}_{OD{C_{inv}}},
\end{aligned}\label{loss3}
\end{array}
\end{equation}
where $\theta _{PDC}$, $\theta _{ODC}$ and $\theta _{DS}$ denote the network parameters of PDC, ODC and DS, respectively. During the training stage, the parameters of the three models are updated in turns by minimizing Eq. (\mbox{\ref{loss1}}), Eq. (\mbox{\ref{loss2}}) and Eq. (\mbox{\ref{loss3}}). To be specific, a) fix $\theta _{DS}$, simultaneously update $\theta _{PDC}$ and $\theta _{ODC}$ by optimizing Eq. (\mbox{\ref{loss1}}) and (\mbox{\ref{loss2}}); b) fix $\theta _{PDC}$ and $\theta _{ODC}$, simultaneously update $\theta _{DS}$ by optimizing Eq. (\mbox{\ref{loss3}}).

\subsection{Network Architecture}

In this section, the connections of the three models (DS, PDC and ODC) and data flow are described in Full Model. In DS, SSD-512 is attach to the 12-th convolutional layer (namely ``conv4\_4$^*$'' layer) of VGG-19. It receives the 512-channel feature map with the 1/16 size of the original input. FCN-8s integrates the outputs of conv3\_4$^*$, conv4\_4$^*$ and conv5\_4$^*$ layers to predict the final segmentation map. In order to obtain a better performance for segmentation, some feature maps from two streams in DS are concatenated at the channel axis, which have the same height and width size. To be specific, the conv5\_4$^*$'s output and conv6\_2\dag's output are concatenated together. Note that ``*'' represents the layer name is from VGG-19 Network \footnote{https://gist.github.com/ksimonyan/3785162f95cd2d5fee77}, and ``\dag'' denotes the layer name comes from SSD Network \footnote{https://github.com/weiliu89/caffe/blob/ssd/examples/ssd/ssd\_pascal.py}.

As for two discriminators, PDC's input is from the feature map of conv5\_4$^*$ layer, and ODC receives the pooled features by ROI pooling operation.

\section{Experiments}
%-------------------------------------------------------------------------
In this section, we respectively report the experimental details and the results of our proposed models, and compare with some existing methods for the same problem.   

\subsection{Datasets}
In order to evaluate our methods, the two popular synthetic datasets are selected: GTA5 \cite{Richter_2016_ECCV} and SYNTHIA \cite{RosCVPR16} as the source domain and choose the Cityscapes \cite{Cordts2016Cityscapes} as the target domain.

\textbf{GTA5} is collected from \emph{ Grand Theft Auto V}, which is a realistic open-world computer game developed by \emph{Rockstar Games}. It contains 24,996 scenes with image size of $1914 \times 1052$ (other abnormal resolution is $1914 \times 1046$) pixels. All scenes are generated from a fictional city of Los Santos in the game, which are based on Los Angeles in Southern California. The annotation classes are compatible with two main datasets: Cityscapes and CamVid \cite{DBLP:journals/prl/BrostowFC09}. In the experiments, the target domain is Cityscapes, so we choose the 19-class ground truth.

%(SYNTHetic collection of Imagery and Annotations)
\textbf{SYNTHIA} is SYNTHetic collection of Imagery and Annotations, a large-scale collection of photo-realistic frames rendered from some virtual cities, which contains 2 image datasets and 7 video sequences, with a resolution $1280 \times 760$. In this paper, we use a subnet of SYNTHIA, called SYNTHIA-RAND-CITYSCAPES as the source domain, of which label space is compatible with the Cityscapes. To be specific, the subnet contains 9,400 images with 13-class categories.

\textbf{Cityscapes} is a real-world urban scenes dataset, which are collected from 50 European cities. In the dataset, about 5,000 images, with high resolution $2018 \times 1024$, are fine annotated at pixel level, which are divided into three subnets with numbers 2,975, 500 and 1,525 for training, validation and testing. It defines 19 common object categories in urban scenes for semantic segmentation. In this paper, all models are tested on the Cityscapes val dataset. 

\textbf{Bounding-box labels} The above three datasets do not provide the object-level annotations. Thus, we need to generate them for DS and ODC. To be specific, the bounding boxes of background objects (sky, building, road, etc.) are collected by transforming from the pixel-wise ground truth. As for the foreground objects (such as pedestrian, bike, car, and so on), only from the per-pixel labels, the bounding boxes of some occluded objects cannot be accurately generated. Therefore, a powerful detection model is adopted, DSOD-300 \mbox{\cite{Shen2017DSOD}} to detect the foreground objects, which is trained on PASCAL VOC 2007 detection dataset. 

\subsection{Evaluation and experimental setup}

\subsubsection{Evaluation}
In the semantic segmentation field, a main metric is Intersection-over-Union (IoU), which is firstly proposed in PASCAL VOC \cite{everingham2015pascal}. Concretely, 
\begin{equation}
\begin{array}{l}
\begin{aligned}
IoU = \dfrac{{TP}}{{TP + FP + FN}},
\end{aligned}\label{IoU}.
\end{array}
\end{equation}
where TP, FP and FN are the numbers of true positive, false positive, and false negative pixels, respectively. 

\subsubsection{Experimental setup}

\textbf{Implementation Details:} In DS model, the VGG-19 net \cite{simonyan2014very} is adopted as the basic neural network. Base on it, the segmentation model is built like the FCN-8s \cite{DBLP:journals/corr/LongSD14}, and the detection model is similar to SSD-512 \cite{liu2016ssd}. The DS network input is the RGB image with size of $512\times 512$ px. During the training stage, the learning rate of basic network is set as ${10^{ - 4}}$, and those of the segmentation and detection streams are set as ${10^{ - 2}}$. PDC and ODC's learning rates are initialized at ${10^{ - 4}}$. DS, PDC and ODC are optimized by SGD, Adam and Adam, respectively. 

\begin{table*}[t]
	\centering
	\caption{Domain adaptation from GTA5 to the Cityscapes val dataset: the comparison results of the mainstream methods and ours. }\label{TableGTA}
	%\setlength\parindent{0em}	
	%\scriptsize
	\renewcommand\arraystretch{1.5}     
	\begin{tabular}{l|p{0.38cm}<{\centering} p{0.38cm}<{\centering} p{0.38cm}<{\centering}  p{0.38cm}<{\centering} p{0.38cm}<{\centering}  p{0.38cm}<{\centering} p{0.38cm}<{\centering} p{0.38cm}<{\centering} p{0.38cm}<{\centering} p{0.38cm}<{\centering} p{0.38cm}<{\centering} p{0.38cm}<{\centering} p{0.38cm}<{\centering} p{0.38cm}<{\centering} p{0.38cm}<{\centering} p{0.38cm}<{\centering} p{0.38cm}<{\centering} p{0.38cm}<{\centering} p{0.38cm}<{\centering} | >{\columncolor{mygray}}p{0.38cm}<{\centering}  }
		\whline
		{Method \%}&\rotatebox{90}{road} &\rotatebox{90}{sidewalk} &\rotatebox{90}{building} &\rotatebox{90}{wall} &\rotatebox{90}{fence} &\rotatebox{90}{pole} &\rotatebox{90}{t light} &\rotatebox{90}{t sign} &\rotatebox{90}{veg} &\rotatebox{90}{terrain} &\rotatebox{90}{sky} &\rotatebox{90}{person} &\rotatebox{90}{rider} &\rotatebox{90}{car} &\rotatebox{90}{truck} &\rotatebox{90}{bus} &\rotatebox{90}{train} &\rotatebox{90}{mbike} &\rotatebox{90}{bike} &\textbf{\rotatebox{90}{mIoU}}\\
		\hline
		{NoAdapt \cite{hoffman2016fcns}} &31.9&18.9&47.7& 7.4& 3.1&16.0&10.4& 1.0&76.5&13.0&58.9&36.0& 1.0&67.1& 9.5& 3.7& 0.0& 0.0& 0.0&21.1\\
		{FCN Wld \cite{hoffman2016fcns}}
		&70.4&32.4&62.1&14.9&5.4&10.9&14.2&2.7&79.2&21.3&64.6&\textbf{44.1}&4.2&70.4&8.0&7.3&0.0&3.5&0.0&27.1\\
		\hline
		{NoAdapt \cite{zhang2017curriculum}} &18.1&6.8&64.1&7.3&8.7&\textbf{21.0}&14.9&\textbf{16.8}&45.9&2.4&64.4&41.6&\textbf{17.5}&55.3&8.4&5.0&\textbf{6.9}&4.3&13.8&22.3  \\
		{CDA \cite{zhang2017curriculum}} &74.9 &22.0 &71.7 &6.0 &11.9 &8.4 &\textbf{16.3} &11.1 &75.7 &13.3 &66.5 &38.0 &9.3 &55.2 &18.8 &18.9 &0.0 &\textbf{16.8} &\textbf{14.6} &28.9 \\
		\hline
		{\textbf{Our methods:}} & & & & & & & & & & & & & & & & & & & & \\
		{DS (NoAdapt)} &65.4&32.4&68.1&14.5&24.8&10.5&4.1&2.0&81.4&34.6&76.5&31.1&0.8&51.6&16.3&8.7&0.0&2.6&0.0&27.7 \\
		{DS+PDC} &71.4&32.6&76.4&28.0&24.9&10.5&4.4&3.8&80.6&29.2&77.4&33.7&1.8&53.6&19.6&18.5&0.0&3.5&0.0&30.0 \\
		{Full} &85.3&43.6&78.5&28.3&25.2&10.5&10.5&6.7&81.4&33.6&74.3&36.7&3.0&73.0&20.2&13.4&0.0&4.7&0.0&33.1 \\
		{Full\dag} &\textbf{89.4}&\textbf{46.4}&\textbf{78.7}&\textbf{34.0}&\textbf{26.9}&15.6&11.8&8.5&\textbf{81.8}&\textbf{40.5}&\textbf{78.6}&36.4&7.3&\textbf{77.9}&\textbf{31.9}&\textbf{33.9}&0.0&8.4&2.4&\textbf{37.4} \\
		\whline
	\end{tabular}
\end{table*}

\begin{figure*}[t]
	\centering
	\includegraphics[width=.98\textwidth]{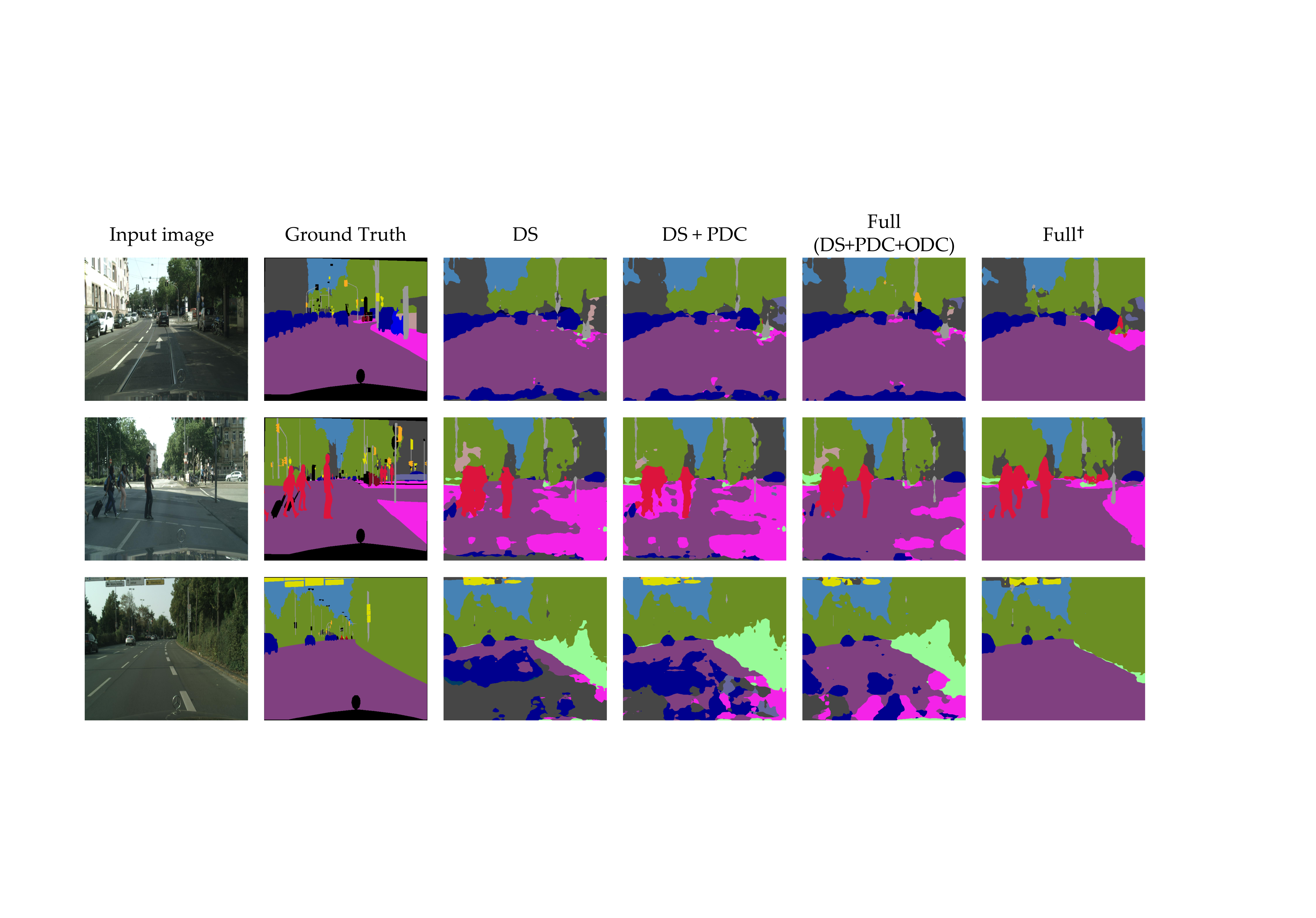}
	\caption{Exemplar results of the Cityscapes val dataset. (Source domain: GTA5) }\label{re_GTA5}
\end{figure*}

\noindent\textbf{Our stepwise experiments. } 

\label{setting}

\begin{itemize}
	\item \textbf{DS: } DS is directly trained without domain adaptation from the source domain to the target domain. %The former has pixel- and object- level labels, but the latter only has object-level labels. 
	
	\item \textbf{DS + PDC: } Based on DS model, PDC is added to the training process by the adversarial learning. 
	
	\item \textbf{Full (DS + PDC + ODC): } In addition to PDC, ODC is also added to the training process by the adversarial learning. 
	
	\item \textbf{Full\dag: } Furthermore, the resnet-152 \cite{he2016deep} is used to initialize the Base Net to verify the proposed method. Other settings are the same as \textbf{Full}.
	
\end{itemize}

\noindent\textbf{Other comparison experiments. }

\begin{itemize}
	
	\item \textbf{FCNs in the wild (FCN Wld) \cite{hoffman2016fcns}: } This work is the first to tackle the same problem as ours. The authors of FCN Wld propose an unsupervised adversarial domain adaptation. Note that the pre-trained model is the dilated VGG-16 \cite{yu2015multi}. 
	
	\item \textbf{CDA \cite{zhang2017curriculum}: } This work is the other existing one to the best of our knowledge. The authors of CDA propose a curriculum-style domain adaptation approach to this problem. For a higher performance, the authors exploit the additional data to train an SVM for superpixel classification. Note that the pre-trained model is the VGG-19, which is the same as ours. 	
\end{itemize}
In the experiments, their no adaptation and final results are listed for comparison with our stepwise experiments.

\begin{table*}[t]
	\centering
	\caption{Domain adaptation from SYNTHIA to the Cityscapes val dataset: the comparison results of the mainstream methods and ours. }\label{TableSYN}
	%\setlength\parindent{0em}	
	%\scriptsize
	\renewcommand\arraystretch{1.5}      
	\begin{tabular}{l | p{0.50cm}<{\centering} p{0.50cm}<{\centering} p{0.50cm}<{\centering}  p{0.50cm}<{\centering} p{0.50cm}<{\centering} p{0.50cm}<{\centering} p{0.50cm}<{\centering} p{0.50cm}<{\centering} p{0.50cm}<{\centering} p{0.50cm}<{\centering} p{0.50cm}<{\centering}  p{0.50cm}<{\centering} p{0.50cm}<{\centering} p{0.50cm}<{\centering} p{0.50cm}<{\centering} p{0.50cm}<{\centering} | >{\columncolor{mygray}}p{0.50cm}<{\centering}  }
		\whline
		{Method \%} &\rotatebox{90}{road} &\rotatebox{90}{sidewalk} &\rotatebox{90}{building} &\rotatebox{90}{wall} &\rotatebox{90}{fence} &\rotatebox{90}{pole} &\rotatebox{90}{t light} &\rotatebox{90}{t sign} &\rotatebox{90}{veg} &\rotatebox{90}{sky} &\rotatebox{90}{person} &\rotatebox{90}{rider} &\rotatebox{90}{car} &\rotatebox{90}{bus}  &\rotatebox{90}{mbike} &\rotatebox{90}{bike} &\textbf{\rotatebox{90}{mIoU}}\\
		\hline
		{NoAdapt \cite{hoffman2016fcns}} &6.4&17.7&29.7&1.2&0.0&15.1&0.0&7.2&30.3&66.8&51.1&1.5&47.3&3.9&0.1&0.0&17.4\\
		{FCN Wld \cite{hoffman2016fcns}} &11.5&19.6&30.8&4.4&0.0&20.3&0.1&\textbf{11.7}&42.3&68.7&\textbf{51.2}&3.8&54.0&3.2&0.2&0.6&20.2\\
		\hline
		{NoAdapt \cite{zhang2017curriculum}} &5.6&11.2&59.6&0.8&0.5&\textbf{21.5}&\textbf{8.0}&5.3&72.4&75.6&35.1&\textbf{9.0}&0.0&0.0&0.5&18.0 &22.0 \\
		{CDA \cite{zhang2017curriculum}}  &65.2 &26.1 &74.9 &0.1 &0.5 &10.7 &3.7 &3.0 &76.1  &70.6 &47.1 &8.2 &43.2 &20.7 &0.7 &13.1 &29.0 \\
		\hline
		{\textbf{Our methods:}}& & & & & & & & & & & & & & & & &  \\
		{DS (NoAdapt)} &52.8&24.2&66.9&6.2&0.0&7.5&0.0&0.0&79.5&75.8&37.8&4.7&64.2&19.2&0.6&16.3&28.5\\
		{DS+PDC} &71.7&34.6&74.6&11.0&0.2&11.6&0.0&2.9&79.9&78.6&39.7&8.6&55.3&20.5&0.9&13.7&31.4 \\
		{Full} &87.4&43.4&\textbf{78.0}&\textbf{16.8}&\textbf{1.8}&11.7&0.0&2.9&\textbf{80.1}&80.5&38.1&8.1&0.0&26.2&\textbf{1.4}&\textbf{19.7}&\textbf{35.7} \\
		{Full\dag} &\textbf{90.2}&\textbf{50.2}&76.6&15.9&0.1&8.6&0.0&1.2&76.8&\textbf{82.6}&36.9&7.1&\textbf{76.7}&\textbf{30.2}&0.0&8.3&35.2 \\
		\whline
	\end{tabular}
\end{table*}

\subsection{GTA5 $\rightarrow$ Cityscapes}
\label{exp_GTA}
Table \ref{TableGTA} lists the qualitative results of some methods for the shift from GTA5 to Cityscapes, including FCN Wld \cite{hoffman2016fcns}, CDA \cite{zhang2017curriculum} and our stepwise experiments: DS, DS+PDC, Full and Full\dag. The bold fonts represent the best of the corresponding column.

From the final results, we can see Full$\dag$ model achieves the best result: mean IoU of 37.4\%. Based on the pre-trained models with the similar learning ability, the mean IoU of Full model (33.1\%) also outperforms that of FCN Wld (27.1\%) and CDA (28.9\%). 
As for the results of the three methods with no adaptation, DS improves the mean IoU (from 21.1/22.3\% to 27.7\%, increasing by 31.3/24.2\%, respectively) significantly. Concretely, the performances for almost all of the categories increase remarkably, which shows the effectiveness of exploiting the bounding-box labels for semantic segmentation. At the same time, it also confirms our observation that object-level features are more robust than local pixel-level features in the cross-domain problem. According to the results of DS+PDC and Full, ODC plays a more important role in learning domain-invariant features than PDC (improvement of 3.1\% versus 2.3\%).

In order to analyze the semantic segmentation performance further and intuitively, Figure \ref{re_GTA5} shows the visualization results of our step-by-step methods. The images in the first column are selected from the Cityscapes val dataset. The second column shows the ground truth, and the remaining columns illustrate the predicted labels of DS, DS+ODC, Full and Full$\dag$ in turns. On the whole, After considering PDC, some segmentation mistakes are removed effectively. From the image in the 2-nd row, introducing the object-level adversarial learning, objects (such as the pedestrian) can be elaborately segmented. Based on the ResNet-152, a better segmentation result can be obtained, which shows the powerful feature learning ability of the residual network.

\begin{figure*}[t]
	\centering
	\includegraphics[width=.98\textwidth]{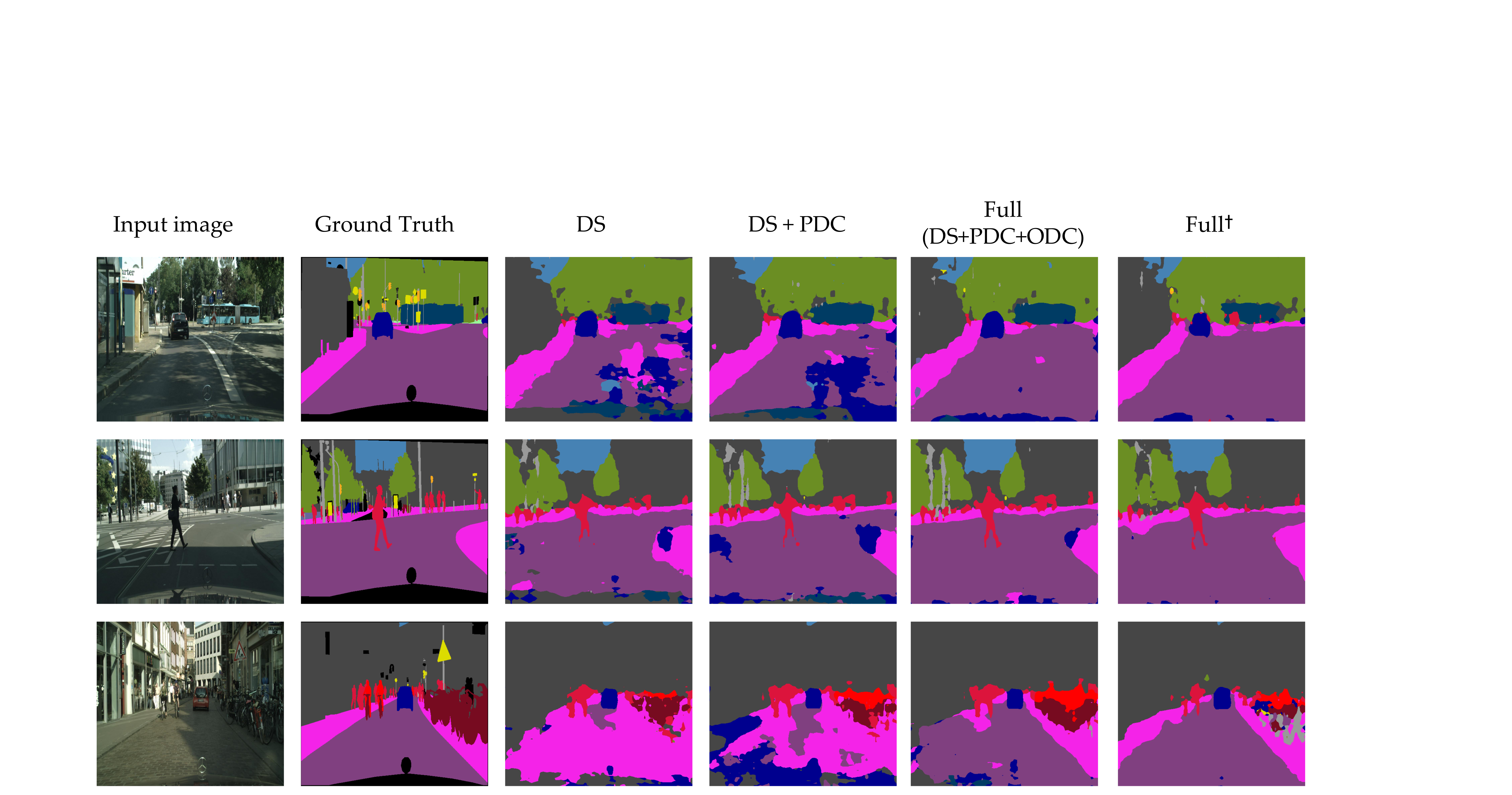}
	\caption{Exemplar results of the Cityscapes val dataset. (Source domain: SYNTHIA) }\label{re_SYN}
\end{figure*}

\subsection{SYNTHIA $\rightarrow$ Cityscapes}

The results of FCN Wld \cite{hoffman2016fcns}, CDA \cite{zhang2017curriculum} and our stepwise experiments (DS, DS+PDC, Full and Full\dag) are listed in Table \ref{TableSYN}, which are adapted from SYNTHIA to the Cityscapes. The bold fonts represent the best of the corresponding column. For a fair comparison with CDA \cite{zhang2017curriculum}, it is noted that the IoU performances of the three items (terrain, truck and train) are removed. This is because the three kinds of objects are not annotated in the source domain: SYNTHIA.  

Similar to Section \ref{exp_GTA}, the proposed method obtains the best performance (35.7\%). Compared with the previous best method CDA (29.0\%), the result of Full contributes 6.7\% raw and 23.1\% relative mean IoU improvement. For the three no adaptation methods, our method prompt many objects' results greatly. In particularly, sky's segmentation IoU achieves the improvement from $\sim6\%$ to $\sim53\%$. According to the mean IoU of DS+PDC and Full, ODC's contribution (4.3\%) is greater than PDC's (2.9\%). Compared with Full, we find that the result of Full$\dag$ has the slight reduction (from 35.7\% to 35.2\%) . The main reason may be the domain gap between SYNTHIA and Cityscapes is larger than that between GTA5 and Cityscapes. Although Full$\dag$ is initialized at ResNet-152, some domain gaps cannot be reduced effectively.

For reporting the advantages of our algorithms, Figure \ref{re_SYN} shows three typical exemplar labeling results. From the visualization results of Column 5 and 6, there is little difference between Full and Full$\dagger$. Some objects' performances of Full$\dag$ are worse than that of Full, such as the bus in Row 1 and the bicycle in Row 3. The other columns show the similar phenomenon to Figure \ref{re_GTA5}.

\subsection{Ablation Study for Bounding-box Labels}

In this paper, the proposed approach exploits the bounding-box labels to train an object detector. Another treatment is: bounding-box labels can be mapped into a coarse segmentation mask, which will also promote to train a coarse semantic segmentor. For a further comparison between the object detector and the coarse segmentor, three groups of experiments with single FCN model (FCN-8s) are conducted: 
\begin{itemize}
\item 1) training with only bounding-box labels on Cityscapes training set; 
\item 2) training with only per-pixel labels on GTA5 and bounding-box labels on Cityscapes; 
\item 3) training with only per-pixel labels on SYNTHIA and bounding-box labels on Cityscapes.
\end{itemize}

\begin{table*}[htbp]
	
	\centering
	\caption{The comparison results of three groups of ablation experiments and our proposed DS/Full model. }
	
	\begin{tabular}{c|cc|c|cc|c}
		\hline
		\multirow{2}{*}{Methods} &\multicolumn{2}{c|}{Domain} &\multirow{2}{*}{Adaptation} 	&\multicolumn{2}{c|}{Labels}		 &\multirow{2}{*}{mean IoU}			\\
		\cline{2-3} \cline{5-6} 
		&source &target &  &bbx &per-pixel	& 		\\
		\hline
		single FCN \#1-1    &\multicolumn{2}{c|}{City}    &\xmark &City &\xmark &23.0	\\
		%single FCN \#1-2   &\multicolumn{2}{c|}{City}   &\checkmark &City &\xmark &23.1	\\
		\hline
		single FCN \#2-1 &GTA5    & City    &\xmark &src \& tgt &src &22.3	\\
		single FCN \#2-2 &GTA5    & City   &\checkmark  &src \& tgt &src  &28.9	\\
		DS \#2   &GTA5      &   City   &\xmark &src \& tgt &src &27.7	\\
		Full \#2   &GTA5      &   City   &\checkmark &src \& tgt &src &33.1	\\
		\hline
		single FCN \#3-1&SYN    & City   &\xmark &src \& tgt &src &19.4	\\
		single FCN \#3-2&SYN    &  City  &\checkmark &src \& tgt &src &24.2	\\
		DS \#3&SYN        &City    &\xmark &src \& tgt &src &28.5	\\
		Full \#3   &SYN      &   City    &\checkmark &src \& tgt &src &35.7	\\
		\hline
		
	\end{tabular}\label{Table-ablation}
\end{table*}

\begin{figure}[t]
	\centering
	\includegraphics[width=.48\textwidth]{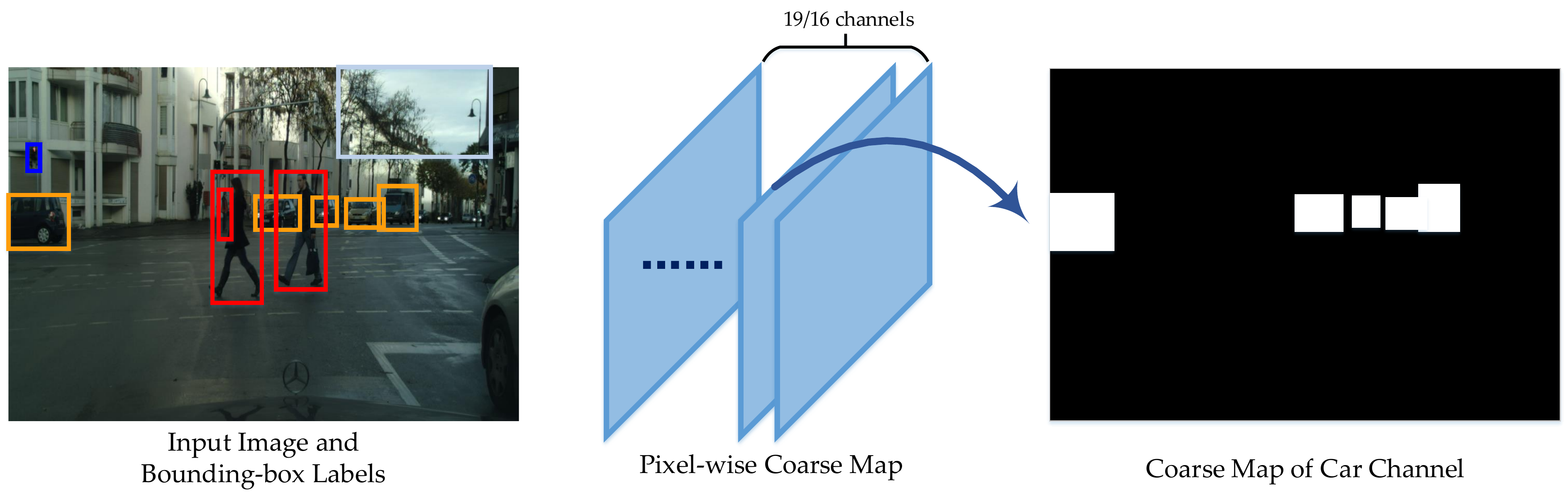}
	\caption{The demonstration of the coarse map.}\label{coarse_map}
\end{figure}

The evaluation of the above experiments is on the val set of Cityscapes. In the three experiments, it is noted that the bounding-box labels represent the coarse maps generated by the bounding-box labels. Specifically, the generation process of coarse maps is explained as below: it is a 19- or 16-channel tensor corresponding to the number of categories in two adaptation experiments (namely GTA5 $\rightarrow$ Cityscapes and SYNTHIA $\rightarrow$ Cityscapes). Each channel with size of input image is a mask for the corresponding category. Figure \mbox{\ref{coarse_map}} illustrates the generation process.

For the first experiment, semantic segmentation is a single-label task (each pixel only has a single label), which outputs the exclusive result. However, because of the overlapped bounding-box labels, the generated rough labels are overlapped. Thus, the above experiments are treated as a multi-label task (each pixel has multiple labels) during the training phase. For comparison with proposed method, the best score from the multi-label outputs are selected as single-label prediction. As for the last two experiments, the FCN has two prediction operations, namely single label on source domain and multi labels on target domain.

Table \ref{Table-ablation} reports the results of the above three groups experiments and the proposed DS model. The DS and Full are our proposed method, which are explained in Section \ref{setting}. ``City'' is shortened form of Cityscapes dataset. From the results of single FCN \#1-x, given the bounding-box labels, the FCN model can learn coarse features to classify each. However, the performance is poor because of some noises in the labels. In the second group of experiment, the DS and Full model respectively outperform the single FCN \#2-1 and \#2-2. Compared with the single FCN, DS exploits the bounding-box labels to train a detector on the two domains. Thus, DS can extract the structured inter- and inter-object features, which are more domain-invariant than local features via the single FCN. Besides, in the adaptation experiments, Full model is trained by the hierarchical (pixel- and object- level) adversarial leaning, which is more robust than only pixel-level adversarial learning in the single FCN. The same phenomenon is shown as in the third experiments.

In summary, the proposed method that trains a detector is more effective than the single FCN in the domain adaption.

\subsection{DS vs. Mask RCNN}

From the aspect of the two paper's purpose, Mask RCNN \mbox{\cite{he2017mask}} is a supervised method for instance segmentation, which does not segment the background objects. For the aspect of architecture, Mask RCNN must detect the objects first and then segment them. Our model consists of two streams, which is an asymmetric multi-task learning on the two domains. In other word, the detection result of Mask RCNN is essential while ours is auxiliary in the test stage.

Even if there are differences, Mask RCNN and DS are both multi-task learning framework for object detection and segmentation. Thus, we conduct two groups of no adaptation experiments using the two algorithms. To be specific, train DS and Mask RCNN on synthetic dataset and test them on the real data. In the Mask RCNN experiments, we implement the code from maskrcnn-benchmark \mbox{\cite{massa2018mrcnn}}. Table \mbox{\ref{maskrcnn}} reports the results of two groups of experiments. From it, Mask RCNN outperforms the proposed DS. We think the main reason is that they have different architectures. As for the two multi-task schemes, Mask RCNN is a sequential architecture, of which segmentation module directly exploits the features from detection. DS is an asymmetric multi-task architecture, of which detection and segmentation modules only share the base features from the backbone. In general, the Mask RCNN's sequential architecture is better than the asymmetric architecture. However, during the test phase, the latter does not need the detection but the former must firstly detect the bounding box. In terms of runtime, DS is faster than Mask RCNN.

\begin{table}[htbp]
	
	\centering
	\caption{The comparison results of DS and Mask RCNN. }
	\begin{tabular}{c|cc|c}
		\hline
		\multirow{2}{*}{Methods} &\multicolumn{2}{c|}{Domain} 	 &\multirow{2}{*}{mean IoU}			\\
		\cline{2-3} 
		&source &target  	& 		\\
		\hline
		DS   &GTA5      &   City     &27.7	\\
		Mask RCNN \cite{he2017mask}  &GTA5      &   City    &\textbf{29.3}	\\
		\hline
		DS &SYN        &City     &28.5	\\
		Mask RCNN \cite{he2017mask}  &SYN      &   City   &\textbf{30.1}	\\
		\hline	
	\end{tabular}	
	\label{maskrcnn}
\end{table}

\subsection{2 $\times$ N-class ODC vs. 2-class ODC}

Traditional domain discriminator only classifies the features' sources, which is a binary classification. In our ODC, we attempt to make it learn the object label and source label of each feature simultaneously. The proposed 2$\times$N-class ODC contains more neural units in fully-connected layer than traditional 2-class ODC. Note that N denotes the number of categories in the dataset. By the supervised training, some specific units of the 2$\times$N-class ODC strongly respond to the specific object category. Thus, the adversarial loss of the specific category cannot suffer from the effects of the other categories. For the 2-class ODC, due to lack of the supervision at the object level, it cannot learn the above ability of the 2$\times$N-class ODC. In summary, the 2$\times$N-class ODC provides more accurate loss than the 2-class ODC. Table \mbox{\ref{com-odc}} reports the results of the full models with the 2$\times$N-class ODC and the 2-class ODC. From it, we find the mIoU of the former is better than that of the latter. 

\begin{table}[htbp]
	
	\centering
	\caption{The comparison results of the Full models with the  2-class ODC and the 2$\times$N-class ODC. }
	\begin{tabular}{c|cc|c}
		\hline
		\multirow{2}{*}{Methods} &\multicolumn{2}{c|}{Domain} 	 &\multirow{2}{*}{mean IoU}			\\
		\cline{2-3} 
		&source &target  	& 		\\
		\hline
		2-class ODC   &GTA5      &   City     &31.5	\\
		2$\times$N-class ODC  &GTA5      &   City    &\textbf{33.1}	\\
		\hline
		2-class ODC &SYN        &City     &34.8	\\
		2$\times$N-class ODC  &SYN      &   City   &\textbf{35.7}	\\
		\hline	
	\end{tabular}	
	\label{com-odc}
\end{table}

\section{Conclusion}

In this paper, we propose a weakly supervised adversarial domain adaptation to improve the segmentation performance from synthetic data to real-world data. To be specific, a weakly supervised model for object detection and semantic segmentation is built, name as DS model, which extract more robust domain-invariant features than the traditional FCN-based methods. In addition, the pixel-/object- level domain classifiers are designed to guide the DS model to learn domain-invariant features by the adversarial learning, which can reduce the domain gap effectively. Our method outperforms all the existing method that do domain adaptation from synthetic scenes to real-world urban scenes for semantic segmentation. In the future work, we will further explore the object relations in the scenes, which is a key domain-invariant feature in the cross-domain semantic segmentation. 

%\section*{ACKNOWLEDGMENT}

\bibliographystyle{IEEEtran}
\bibliography{IEEEabrv,reference}

\newpage

\begin{IEEEbiography}[{\includegraphics[width=1in,height=1.25in,clip,keepaspectratio]{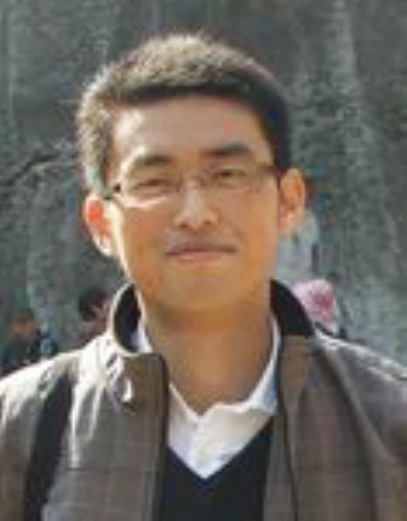}}]{Qi Wang} (M'15-SM'15) received the B.E. degree in automation and the Ph.D. degree in pattern recognition and intelligent systems from the University of Science and Technology of China, Hefei, China, in 2005  and 2010, respectively.  He is currently a Professor with the School of Computer Science and with the Center for OPTical IMagery Analysis and Learning (OPTIMAL), Northwestern Polytechnical University, Xi'an, China. His research interests include computer vision and pattern recognition.
\end{IEEEbiography}

\begin{IEEEbiography}[{\includegraphics[width=1in,height=1.25in,clip,keepaspectratio]{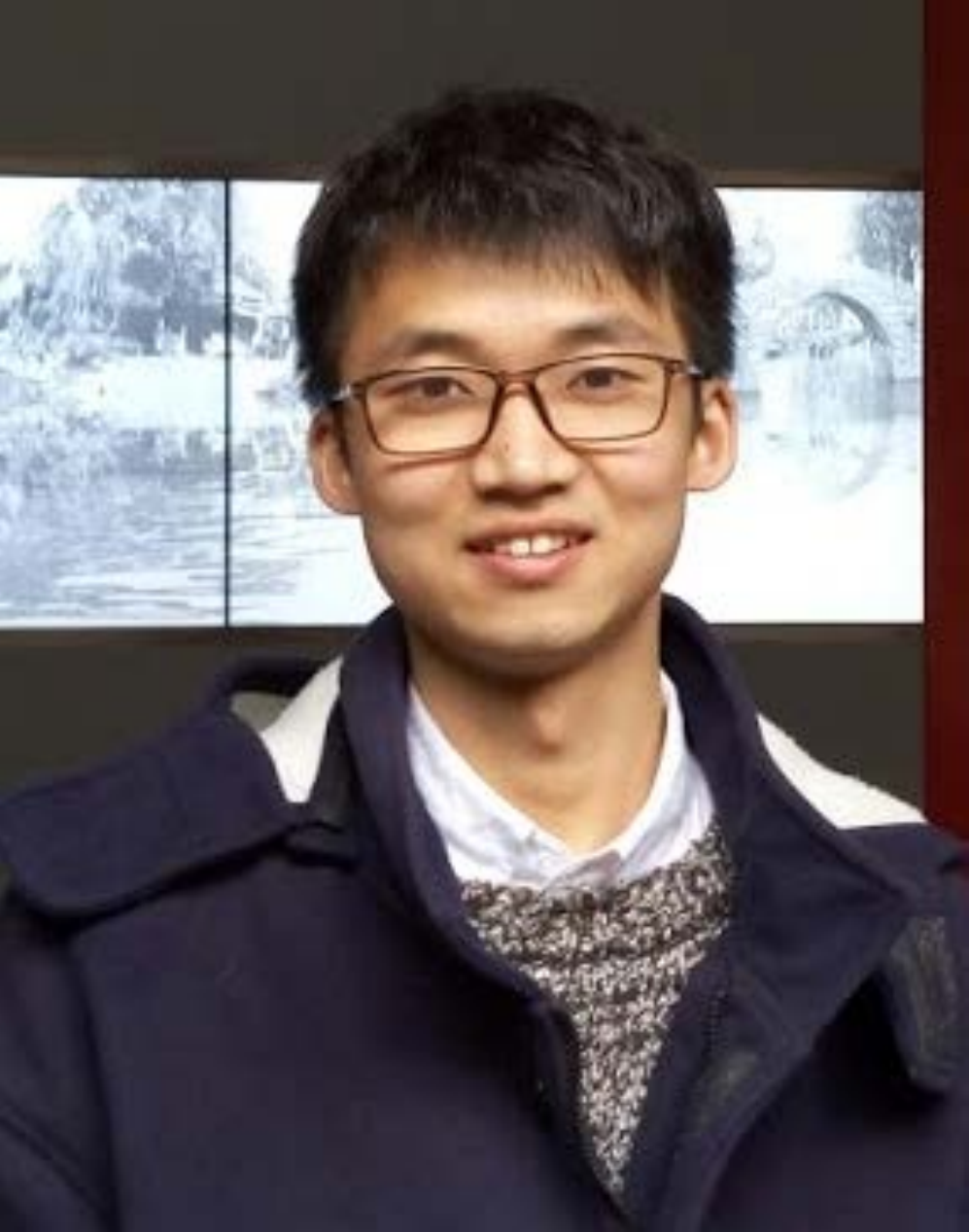}}]{Junyu Gao} received the B.E. degree in computer science and technology from the Northwestern Polytechnical University, Xi'an 710072, Shaanxi, P. R. China, in 2015. He is currently pursuing the Ph.D. degree from Center for Optical Imagery Analysis and Learning, Northwestern Polytechnical University, Xi'an, China. His research interests include computer vision and pattern recognition.
\end{IEEEbiography}

\begin{IEEEbiographynophoto}{Xuelong Li} (M'02-SM'07-F'12) is a full professor with the School of Computer Science and the Center for OPTical IMagery Analysis and Learning (OPTIMAL), Northwestern Polytechnical University, Xi'an 710072, Shaanxi, P. R. China.
\end{IEEEbiographynophoto}

\end{document}